\documentclass[11pt, a4paper]{article}
\usepackage{vbarticle}
\backreftype{none}
\usepackage{booktabs}
\usepackage{tipa}
\usepackage{soul}
\usepackage{tikz-dependency}
\tikzset{/depgraph/.cd, /depgraph/.search also={/tikz}, theme=simple, label style={font=\large}, edge style={thick}}

\usepackage[T5,T1]{fontenc}
\DeclareTextSymbolDefault{\ohorn}{T5}
\DeclareTextSymbolDefault{\uhorn}{T5}

\newcommand*{\theaffiliation}{}
\newcommand*{\affiliation}[1]{\renewcommand{\theaffiliation}{#1}}

\makeatletter
\renewcommand\maketitle{
{\raggedright
\begin{center}
{\marginnote{{\sffamily\large \textsc{\MakeLowercase{\thesubtitle}}}} \LARGE\sffamily \@title }\\[4ex] 
{\marginnote{\@date} \large \@author}\\
\theaffiliation
\end{center}}}
\makeatother

\title{MaiBaam Annotation Guidelines}
\subtitle{Bavarian UPOS \& UD}
\date{October 31, 2025\\
Guidelines version 1.2\\UD release 2.17}
\author{Verena Blaschke, Barbara Kova\v{c}i\'{c}, Siyao Peng, Barbara Plank}
\affiliation{MaiNLP, CIS, LMU Munich\\\texttt{verena.blaschke@cis.lmu.de}}

\usepackage{ulem}
\usepackage{contour}

\contourlength{0.8pt}
\renewcommand{\ul}[1]{%
  \uline{\phantom{#1}}%
  \llap{\contour{white}{#1}}%
}

\newcommand*{\pos}[1]{{\footnotesize\textls{\texttt{\MakeUppercase{#1}}}}}
\newcommand*{\tag}[2]{\textit{#1}\kern1pt\textsubscript{\pos{#2}}}
\newcommand*{\dep}[1]{{\sffamily #1}}
\newcommand*{\dependent}[1]{\textbf{#1}}
\newcommand*{\head}[1]{\ul{#1}}
\newcommand*{\leftdep}[3]{#1~$\leftarrow$\kern-1pt#2\kern-1pt---~#3}
\newcommand*{\rightdep}[3]{#1~---\kern-1pt#2\kern-1pt$\rightarrow$~#3}
\newcommand*{\gloss}[1]{#1}

\newcommand{\amod}{\dep{amod}}
\newcommand{\acl}{\dep{acl}}
\newcommand{\aclrelcl}{\dep{acl:relcl}}
\newcommand{\advcl}{\dep{advcl}}
\newcommand{\advmod}{\dep{advmod}}
\newcommand{\appos}{\dep{appos}}
\newcommand{\aux}{\dep{aux}}
\newcommand{\case}{\dep{case}}
\newcommand{\cc}{\dep{cc}}
\newcommand{\ccomp}{\dep{ccomp}}
\newcommand{\compound}{\dep{compound}}
\newcommand{\conj}{\dep{conj}}

\newcommand{\dett}{\dep{det}}
\newcommand{\detposs}{\dep{det:poss}}
\newcommand{\flatt}{\dep{flat}}
\newcommand{\fixed}{\dep{fixed}}
\newcommand{\goeswith}{\dep{goeswith}}
\newcommand{\iobj}{\dep{iobj}}
\newcommand{\markk}{\dep{mark}}
\newcommand{\nmod}{\dep{nmod}}
\newcommand{\nsubj}{\dep{nsubj}}
\newcommand{\obj}{\dep{obj}}
\newcommand{\obl}{\dep{obl}}
\newcommand{\oblarg}{\dep{obl:arg}}
\newcommand{\parataxis}{\dep{parataxis}}
\newcommand{\punct}{\dep{punct}}
\newcommand{\vocative}{\dep{vocative}}
\newcommand{\xcomp}{\dep{xcomp}}

\newcommand*{\udlink}[2]{\href{#2}{\textsc{\MakeLowercase{#1}}}}
\newcommand*{\ulink}[1]{(\udlink{u}{#1})}
\newcommand*{\delink}[1]{(\udlink{de}{#1})}
\newcommand*{\udelink}[2]{(\udlink{u}{#1}, \udlink{de}{#2})}

\newcommand*{\rightarr}{$\rightarrow$}

\newcommand*{\sent}[2] %
{\textit{#2} (#1)}

\definecolor{lightteal}{RGB}{145, 219, 219}
\definecolor{lightyellow}{RGB}{240, 240, 137}
\newcommand*{\datlikeacc}[1]{\sethlcolor{lightteal}\hl{#1}}
\newcommand*{\acclikedat}[1]{\sethlcolor{lightyellow}\hl{#1}}
\newcommand*{\clitic}[1]{\sethlcolor{lightteal}\hl{#1}}
\newcommand*{\fullpron}[1]{\sethlcolor{lightyellow}\hl{#1}}

\newcommand*{\fslash}{/\allowbreak{}}

\usepackage{fontawesome}
\usepackage{tcolorbox}
\newenvironment{warning}[1]
{\begin{tcolorbox}[colback=white,colframe=red!20]
\noindent \textcolor{red!40}{\faExclamationTriangle{} #1}
}
{ 
\end{tcolorbox}
}

\usepackage{enumitem}
\setlist[1]{itemsep=0pt, topsep=3pt}
\setlist[2]{itemsep=0pt, topsep=0pt}

\newcommand*{\translation}[1]{\textcolor{black!50}{`#1'}}
\newcommand*{\translationwithnote}[2]{\textcolor{black!50}{#1~`#2'}}
\newcommand*{\attribution}[1]{\phantom{..}\hfill(#1)}

\newlength{\spacer}
\newcommand{\margintext}[1]{{\sffamily\small #1}}

\begin{document}

\maketitle

\pagenumbering{roman}
\tableofcontents

\newpage
\pagenumbering{arabic}

\section*{General remarks}
\label{sec:remarks}
\addcontentsline{toc}{section}{General remarks}
This document provides annotation guidelines for MaiBaam \citep{blaschke2024maibaam}, a Bavarian corpus manually annotated with part-of-speech (POS) tags, syntactic dependencies, and German lemmas.
MaiBaam belongs to the Universal Dependencies (UD) project \citep{zeman2023ud-2-12, demarneffe2021ud}, and our annotations elaborate on the general and German UD version~2 guidelines.

This document is structured broadly in the order we prepare and annotate sentences: first, preprocessing and tokenization~(\S\ref{sec:prep-tok}), then general recaps of POS tags~(\S\ref{sec:pos}) and dependencies~(\S\ref{sec:dep}), then a note on adding German lemmas~(\S\ref{sec:lemmas}), before we go into annotation decisions that would also apply to German~(\S\ref{sec:general}) and lastly decisions that are specific to Bavarian grammar~(\S\ref{sec:bavarian}).

Many examples are written in German, since the standardized orthography makes it easier to search this PDF.
We annotate UD-style POS tags (UPOS tags) and dependencies, add German lemmas, and add features related to whitespace and typographical errors where appropriate, but do not add any other information (no Bavarian lemma, XPOS tags, morphological features, enhanced dependencies or miscellaneous annotations).

This document is primarily directed at present and future annotators of MaiBaam.
We publish it to additionally allow others working with MaiBaam or annotating similar data to better understand the decisions we have made.
These rules are not set in stone.
If you are a MaiBaam annotator and annotating something and applying one of the rules here would make for an awkward and unintuitive annotation, please bring it up for discussion.
Likewise, if you are unsure about how to annotate a word\fslash{}phrase, please also raise it as a discussion point.\\

\noindent
We use the following notation in this document:
\begin{itemize}
    \item Part-of-speech tags: \pos{tag} in small caps; \tag{tagged word in italics}{tag}
    \item Dependency relations: \dep{name} of dependency relation in sans-serif, \head{head} word underlined and \dependent{dependent} in boldface
    \item Where possible, we use arrows to show dependencies:\\ \rightdep{head}{\dep{dependency}}{dependent}
\end{itemize}

\noindent
We reference the following UD treebanks in this document and/or used them for guidance/comparison:
\begin{itemize}
    \item German: \href{https://github.com/UniversalDependencies/UD_German-GSD/}{GSD} \citep{mcdonald-etal-2013-universal}, \href{https://github.com/UniversalDependencies/UD_German-HDT}{HDT} \citep{borges-volker-etal-2019-hdt}, \href{https://github.com/UniversalDependencies/UD_German-PUD/}{PUD} \citep{zeman-etal-2017-conll}, \href{https://github.com/UniversalDependencies/UD_German-LIT}{LIT} \citep{salomoni2017lit}
    \item Swiss German: \href{https://github.com/UniversalDependencies/UD_Swiss_German-UZH}{UZH} \citep{aepli2018parsing}
    \item Low Saxon: \href{https://github.com/UniversalDependencies/UD_Low_Saxon-LSDC}{LSDC} \citep{siewert-etal-2021-towards}
    \item English: \href{https://github.com/UniversalDependencies/UD_English-GUM}{GUM} \citep{zeldes2017gum}, \href{https://github.com/UniversalDependencies/UD_English-EWT}{EWT} \citep{silveira14gold}
\end{itemize}

\subsection*{Changelog}
\addcontentsline{toc}{subsection}{Changelog}

\begin{itemize}
    \item Version 1.2 (2025--10--31): minor clarifications/updates (\S\ref{sec:specific-deprels}~\dep{iobj} and \dep{xcomp}, \S\ref{sec:typos}~typo section, \S\ref{sec:durch-des}~\textit{durch/fir des}, \S\ref{sec:future}~future updates section)
    \item Version 1.1 (2024--10--18): German lemmas added~(\S\ref{sec:lemmas})
    \item Version 1.0 (2024--03--09): First version
\end{itemize}

\newpage
\section{Preprocessing and tokenization}
\label{sec:prep-tok}
\subsection{Preprocessing}
We do not preserve or otherwise mimic the original formatting -- italics, boldface, font size differences, etc. simply disappear.
If there is a case where the original formatting actually is crucial, bring it up for discussion.
In the context of Wikipedia discussion pages, we anonymize usernames by replacing them with USERNAME~(\S\ref{sec:anonymized-names}).

When selecting sentences to annotate, pick full paragraphs if possible (that are then sentence-split).
We skip lists in wiki articles, unless the lists contain full sentences. In that case, include the sentences as individual entries, and skip the bullet points.
Examples for lists to be skipped are the lists of Munich boroughs and neighbouring municipalities \href{https://bar.wikipedia.org/w/index.php?title=Minga&oldid=841494#Mingara_Beziak}{here}.
This is for two reasons: One, they are very long yet structurally not very interesting. Two, for the sentence still to make sense, we would either need to preserve formatting information (line breaks, bullets, even indentation in the case of the list of municipalities) or actually change the sentence by, e.g., adding commas.

We do not correct typos or punctuation errors.
See~\S\ref{sec:typos} for how to annotate typos.

\subsection{Sentence-splitting}
Sentence-splitting is for the most part straight-forward.
Dialogue tags are generally part of the sentence, e.g., the following is a single sentence:\\\textit{The father said: ``Hänsel, go and fetch some wood.''}

\subsection{Metadata}

We include the following sentence-level metadata:
\begin{itemize}
    \item \texttt{sent\_id}: Unique ID that also encodes the source of the sentence.
    \item \texttt{text}: The sentence.
    \item \texttt{text\_en}: The original text (for sentences translated from English).
    \item \texttt{genre}: The text genre. Our genres currently are \textit{wiki} (Wikipedia articles), \textit{social} (Wikipedia discussions), \textit{fiction} (fairy tales), \textit{grammar examples} (Tatoeba sentences, example sentences from Wikipedia pages about grammar, other linguistic example sentences), and \textit{non-fiction} (queries for virtual assistants).
    \item \texttt{dialect\_group}: One of \textit{north, northcentral, central, southcentral, south, unk} or a more precise elaboration on \textit{unk} if possible, e.g., \textit{unk (southcentral\fslash{}south),} with the options sorted from North to South. Use the map on the following page for guidance.
    \item \texttt{location}: The city or municipality if known, else the state or province, else the country, else \textit{unk.} We use English location names.
    \item \texttt{source}: The URL of the Wikipedia or Tatoeba page.
    \item \texttt{author}: The username of a Tatoeba sentence's author.
\end{itemize}

\vspace{\baselineskip}
\marginnote{\margintext{Bavarian dialect areas, based on the classification by \citet[map~47.4]{wiesinger1983einteilung}.}}[-6pt]
\noindent
\includegraphics[width=\textwidth, trim={3mm 0 0 3mm}, clip]{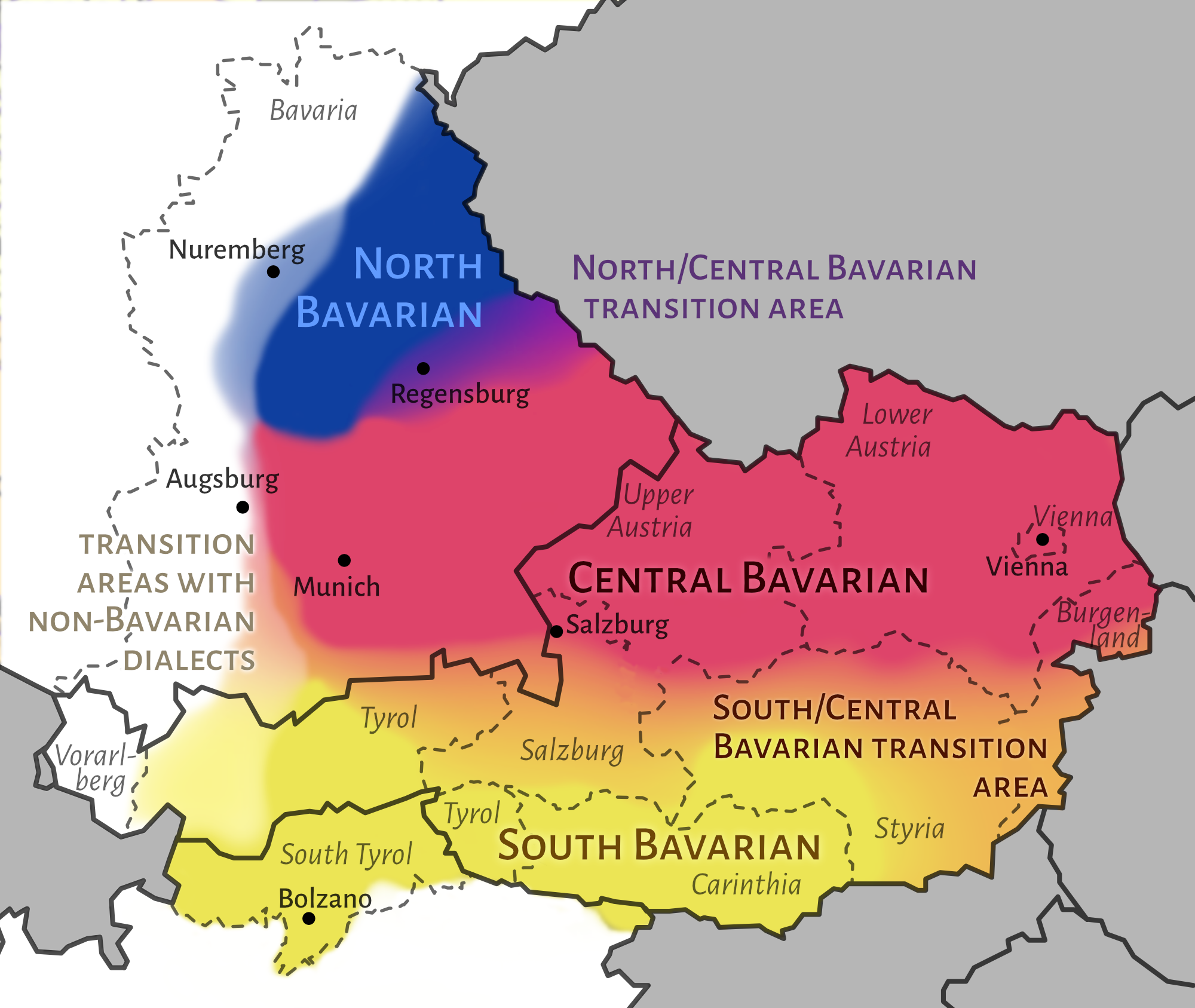}

\subsection{General tokenization guidelines}
\enlargethispage{\baselineskip}

We generally base tokenization decisions on whitespace and punctuation, with the following special cases:
\marginnote{\margintext{\llap{*}This means we follow HDT, but not GSD, PUD or LIT.}}[20pt]

\begin{itemize}
    \item Do not split compound nouns: \textit{Silben-Trennung} is a single token.*
    \item Keep the whitespace-based tokenization for truncated words in cases like \tag{Sonn-}{noun} \tag{und}{cconj} \tag{Feiertage}{noun} \translation{sun- and holidays} (cf.~\S\ref{sec:truncated}).
    \item Split numbers and units: \tag{8}{num} \tag{kg}{noun}
    \item Split up ranges: \tag{400}{num} \tag{--}{adp} \tag{500}{num}
    \item Split off the outer brackets or slashes around phonetic transcriptions, but no other punctation marks inside the transcriptions:\\\tag{[}{punct}~\tag{m\i\ng(\textlengthmark)\textturna}{x}~\tag{]}{punct}
    \item For words erroneously split in the raw data (e.g., \textit{zu mindest} instead of \textit{zumindest} `at least'), follow the instructions in the \href{https://universaldependencies.org/u/dep/goeswith.html}{\goeswith} documentation. See also the note at the end of \S\ref{sec:complementizer-agreement}.
\end{itemize}

\paragraph{Sandhi} 
When a vowel-initial word is appended to a vowel-final word, a linking consonant can be inserted in between \citep[pp.~30--33]{merkle1993bairische}.
In this case, we include the consonant with the first word (e.g., we analyze \textit{wiera} \translation{how he} with its linking \textit{-r-} as \textit{wier} and \textit{a}).

\subsection{Multi-word tokens}
\paragraph{Preposition + determiner} 
We follow the \href{https://universaldependencies.org/de/index.html}{guidelines} for Standard German (see also \citealp{grunewald-friedrich-2020-unifying}) and split fused prepositions and determiners into subtokens.
This was also adopted by the guidelines for \href{https://universaldependencies.org/nds/index.html}{Low Saxon}, but not for \href{https://universaldependencies.org/gsw/index.html}{Swiss German}.
Since different writers use different orthographic styles and since there is variation also in the way determiners are pronounced, we simply split the words into substrings (rather than normalizing them to an arbitrary standard):
\begin{itemize}
    \item \textit{zum} \rightarr{} \tag{zu}{adp} \tag{m}{det} \translation{to the}
    \item \textit{aus'n} \rightarr{} \tag{aus}{adp} \tag{'n}{det} \translation{from the}
\end{itemize}
Sometimes, this can result in slightly awkward tokenizations:
\begin{itemize}
    \item \textit{im} \rightarr{} \tag{i}{adp} \tag{m}{det}  \translation{in the}
\end{itemize}

\paragraph{Particle + determiner (zum)} When \textit{zum} (\textit{zun, zan,} ...) is used in an infinitive construction, we treat it as a multi-word token (\tag{zu}{part} \tag{m}{det}).
For more details, see~\S\ref{sec:infinitive}.

\subsection{Split with \texttt{SpaceAfter=No}}
\label{sec:space-after-no}

\paragraph{Shortened determiners/adpositions in noun phrases}
We also split these off with \texttt{SpaceAfter=No} (instead of MWTs):
\begin{itemize}
    \item \textit{z'Minga} \rightarr{} \tag{z'}{adp} \tag{Minga}{propn}  \translation{in Munich}
    \item \textit{d'neie} \rightarr{} \tag{d'}{det} \tag{neie}{adj}  \translation{the new [one]}
    \item \textit{s´Haus} \rightarr{} \tag{s´}{det} \tag{Haus}{noun}  \translation{the house}
\end{itemize}
\noindent
This is analogous to how shortened determiners and prepositions are treated in French UD treebanks \textit{(d', l').}

\paragraph{Verb/complementizer + pronoun(s)}
In sentences where a verb or conjunction is immediately followed by one or more pronouns, we use \texttt{SpaceAfter=No} to split them:
\begin{itemize}
    \item \textit{gibts} \rightarr{} \tag{gibt}{verb} \tag{s}{pron} \translation{there is}
    \item \textit{håmas} \rightarr{} \tag{hå}{verb} \tag{ma}{pron} \tag{s}{pron} \translation{we have it}
\end{itemize}
\noindent
This is similar to how shortened pronouns are treated in French treebanks (e.g., in \textit{je t'ai vu} \translation{I saw you}, \tag{t'}{pron} and \tag{ai}{aux} get split apart).

\begin{warning}{Exception}
    For conjunctions marked for 2\textsc{sg}, 2\textsc{pl} or 1\textsc{pl}, see \S\ref{sec:complementizer-agreement}.
\end{warning}

\paragraph{Other fused tokens}
In general, use \texttt{SpaceAfter=No}, but feel free to bring up such cases for discussion.
See also~\S\ref{sec:typos}.

\newpage
\enlargethispage{5\baselineskip}
\section{POS tags}
\label{sec:pos}
The detailed general \textsc{(u)} and German \textsc{(de)} guidelines are linked.\\
\begin{tabular}{@{}ll@{\hspace{-30pt}}r@{}}
\toprule
{POS tag} & \multicolumn{2}{l}{Examples \hfill Cases \textit{not} covered by this tag} \\ \midrule
\pos{adj} adjective \ulink{https://universaldependencies.org/u/pos/ADJ.html} & \textit{grün, französisch, zweites,}\\
& \textit{reisend} (present participle used as adjective),\\
& \textit{{[}die{]} 80er {[}Jahre{]}, {[}ein{]} paar} \\[\spacer]
\pos{adp} adposition \ulink{https://universaldependencies.org/u/pos/ADP.html} & \textit{auf, in, bis,}\\
& split-off particles of particle verbs: \textit{[er gibt] auf} \\[\spacer]
\pos{ADV} adverb \ulink{https://universaldependencies.org/u/pos/ADV.html} & \textit{sehr, morgen, hinauf, wann/wie/warum,}  & \tag{nicht}{part}, \\
& \textit{hier, irgendwo, immer, zuerst, zweimal,} & \tag{wer/was}{pron}\\
& \textit{wodurch, gerne, selbst, dazu, gar, nimmer,}\\
&modal particles\\[\spacer]
\pos{AUX} auxiliary \ulink{https://universaldependencies.org/u/pos/AUX.html} & \textit{sollen/müssen/..., würde,} copula \textit{(sein),}  \\
& \textit{haben/sein/werden} indicating tense,\\
& \textit{werden} indicating passive voice &  \\[\spacer]
\pos{CCONJ} coordinating & \textit{und, oder, aber, sondern, weder noch,} & \tag{noch}{adv} \\
conjunction \ulink{https://universaldependencies.org/u/pos/CCONJ.html} & \textit{{[}sowohl...{]} als {[}auch{]}} & (on its own),\\
&  \textit{wie} when not used for interrogation/comparison & \tag{wie}{adv}...?\\[\spacer]
\pos{DET} determiner \udelink{https://universaldependencies.org/u/pos/DET.html}{https://universaldependencies.org/de/pos/DET.html} & articles, possessive pronouns \textit{(ihr/sein/mein)}, & relative pronouns: \\
& \textit{jede, keine, dieselbe, selber, selbig, derjenige,} & \tag{der/die/das}{pron}\\
& \textit{welche, alle, viel(e), wenig(e), weniger, ander(e)}\\
& (also in \pos{det+det} constructions: \textit{die andere [Seite])}\\[\spacer]
\pos{INTJ} interjection \ulink{https://universaldependencies.org/u/pos/INTJ.html}& \textit{hallo, ach, ja} (as a response) &  \textit{ja} as particle (\pos{adv}) \\[\spacer]
\pos{NOUN} noun \ulink{https://universaldependencies.org/u/pos/NOUN.html} & \textit{Letzterer, kg} &  \\[\spacer]
\pos{NUM} numeral \ulink{https://universaldependencies.org/u/pos/NUM.html} & \textit{4, vier} & \tag{viertes/4.}{adj}, \tag{viermal}{adv} \\[\spacer]
\pos{PART} particle \ulink{https://universaldependencies.org/u/pos/PART.html} & \textit{nicht; zu} when used with infinitives & verbal particles: \pos{adp}; \\
&& modal particles: \pos{adv}\\[\spacer]
\pos{PRON} pronoun \udelink{https://universaldependencies.org/u/pos/PRON.html}{https://universaldependencies.org/de/pos/PRON.html} & \textit{ich/mich/mir, wer, jemand, etwas, man,} & \tag{mein/dein/etc.}{det}, \\
& \textit{jedermann, niemand, nichts, sich,}  & \tag{einige}{det}, \tag{irgendein}{det}\\
&relative pronouns\\[\spacer]
\pos{PROPN} proper noun \ulink{https://universaldependencies.org/u/pos/PROPN.html} & \textit{Minga} &{normal tags when possible} \\[\spacer]
\pos{PUNCT} punctuation \ulink{https://universaldependencies.org/u/pos/PUNCT.html} & \textit{. , ? ! /} \\[\spacer]
\pos{SCONJ} subordinating & \textit{dass, falls, damit, weil, wenn} \\
conjunction \ulink{https://universaldependencies.org/u/pos/SCONJ.html} &  &  \\[\spacer]
\pos{SYM} symbol  \ulink{https://universaldependencies.org/u/pos/SYM.html}& \% &  \\[\spacer]
\pos{VERB} verb \ulink{https://universaldependencies.org/u/pos/VERB.html} & \textit{{[}ansässig{]} werden, {[}zu] finden sein, [etw.] haben}  \\[\spacer]
\pos{X} other \ulink{https://universaldependencies.org/u/pos/X.html}& non-words/non-symbols,  & \pos{PROPN} when possible \\ 
& botanical/zoological names\\
\bottomrule
\end{tabular}

\newpage

\newpage
\section{Syntactic dependencies}
\label{sec:dep}
\subsection{Overview}
The detailed general \textsc{(u)} and German \textsc{(de)} guidelines are linked.\\
\renewcommand*{\arraystretch}{1.3}
\begin{tabular}{@{}lll@{}}
\toprule
\multicolumn{3}{l}{\textit{\head{Predicate} \rightarr{} \dependent{noun-like}}} \\
\dep{nsubj} \udelink{https://universaldependencies.org/u/dep/nsubj.html}{https://universaldependencies.org/de/dep/nsubj.html} & subject of verb & \dependent{Sie} \head{schläft} \\
\dep{nsubj:pass} \udelink{https://universaldependencies.org/u/dep/nsubj-pass.html}{https://universaldependencies.org/de/dep/nsubj-pass.html} & subject of passive construction & \dependent{Sie} wurde \head{gesehen} \\
\dep{obj} \udelink{https://universaldependencies.org/u/dep/obj.html}{https://universaldependencies.org/de/dep/obj.html} & accusative object & Ich \head{gebe} dir das \dependent{Bild} \\
\dep{iobj} \udelink{https://universaldependencies.org/u/dep/iobj.html}{https://universaldependencies.org/de/dep/iobj.html}& 2nd accusative object & Das \head{kostet} \dependent{mich} einen Euro \\
\dep{obl} \udelink{https://universaldependencies.org/u/dep/obl.html}{https://universaldependencies.org/de/dep/obl.html} & prepositional phrase & Das Paket \head{liegt} vor der \dependent{Tür} \\
\dep{obl:arg} \udelink{https://universaldependencies.org/u/dep/obl-arg.html}{https://universaldependencies.org/de/dep/obl-arg.html} & dative object & Ich \head{gebe} \dependent{dir} das Bild \\
\dep{obl:agent} \udelink{https://universaldependencies.org/u/dep/obl-agent.html}{https://universaldependencies.org/de/dep/obl-agent.html} & agent of passive construction & Sie wurde von \dependent{mir} \head{gesehen} \\
\dep{expl} \udelink{https://universaldependencies.org/u/dep/expl.html}{https://universaldependencies.org/de/dep/expl.html} & dummy \textit{es} & \dependent{Es} \head{macht} Spaß \\
\dep{expl:pv} \udelink{https://universaldependencies.org/u/dep/expl-pv.html}{https://universaldependencies.org/de/dep/expl-pv.html} & lexicalized reflexive pronouns & Ich \head{bedanke} \dependent{mich} \\
\dep{vocative} \udelink{https://universaldependencies.org/u/dep/vocative.html}{https://universaldependencies.org/de/dep/vocative.html} & addressed listener & \dependent{Marie}, \head{kommst} du mit? \\
\midrule
\multicolumn{3}{l}{\textit{\head{Predicate} \rightarr{} \dependent{predicate}}} \\
\dep{csubj} \udelink{https://universaldependencies.org/u/dep/csubj.html}{https://universaldependencies.org/de/dep/csubj.html} & clausal subject & Ob das \dependent{hilft}, \head{weiß} ich nicht \\
\dep{csubj:pass} \udelink{https://universaldependencies.org/u/dep/csubj-pass.html}{https://universaldependencies.org/de/dep/csubj-pass.html} & clausal subj. of passive clause & Ob das \dependent{hilft}, wurde mir nicht \head{gesagt} \\
\dep{ccomp} \udelink{https://universaldependencies.org/u/dep/ccomp.html}{https://universaldependencies.org/de/dep/ccomp.html} & complement with own subject & Ich \head{wette}, dass du \dependent{gewinnst} \\
\dep{xcomp} \udelink{https://universaldependencies.org/u/dep/xcomp.html}{https://universaldependencies.org/de/dep/xcomp.html} & complement with shared subj. & Ich \head{versuche}, den Text zu \dependent{schreiben} \\
\dep{advcl} \udelink{https://universaldependencies.org/u/dep/advcl.html}{https://universaldependencies.org/de/dep/advcl.html} & adverbial clause & Ich \head{schreibe}, damit sie \dependent{antwortet} \\
\dep{advcl:relcl} \delink{https://universaldependencies.org/de/dep/advcl-relcl.html} & relative clause modifying clause & Sie \head{tanzt}, was ich cool \dependent{finde} \\
\midrule
\multicolumn{3}{l}{\textit{\head{Predicate} \rightarr{} \dependent{auxiliary}}} \\
\dep{aux} \udelink{https://universaldependencies.org/u/dep/aux_.html}{https://universaldependencies.org/de/dep/aux_.html} & auxiliary & Ich \dependent{bin} \head{gegangen} \\
\dep{aux:pass} \udelink{https://universaldependencies.org/u/dep/aux-pass.html}{https://universaldependencies.org/de/dep/aux-pass.html} & passive auxiliary & Ich \dependent{wurde} \head{gesehen} \\
\dep{cop} \udelink{https://universaldependencies.org/u/dep/cop.html}{https://universaldependencies.org/de/dep/cop.html} & copula & Du \dependent{bist} \head{groß} \\
\midrule
\multicolumn{3}{l}{\textit{\head{Predicate} \rightarr{} \dependent{\_\_\_\_}}} \\
\dep{mark} \udelink{https://universaldependencies.org/u/dep/mark.html}{https://universaldependencies.org/de/dep/mark.html} & subordinating conjunction, \textit{zu} & Ich wette, \dependent{dass} du \head{gewinnst} \\
\dep{compound:prt} \udelink{https://universaldependencies.org/u/dep/compound-prt.html}{https://universaldependencies.org/de/dep/compound-prt.html} & particle belonging to verb & Ich \head{fange} \dependent{an} \\
\dep{dislocated} \udelink{https://universaldependencies.org/u/dep/dislocated.html}{https://universaldependencies.org/de/dep/dislocated.html} & dislocated phrase & Das \dependent{Bild}, das \head{gebe} ich dir später \\
\dep{discourse} \ulink{https://universaldependencies.org/u/dep/discourse.html} & interjections, fillers & \dependent{Hey}, \head{kommst} du mit? \\
\midrule
\end{tabular}

{~\hfill ... \textit{continued on the next page}}

\newpage
\textit{Continued from the previous page}

\begin{tabular}{@{}lll@{}}
\midrule
\multicolumn{3}{l}{\textit{\head{Noun-like} \rightarr{} \dependent{noun-like}}} \\
\dep{nmod} \udelink{https://universaldependencies.org/u/dep/nmod.html}{https://universaldependencies.org/de/dep/nmod.html} & noun modifying noun & das \head{Buch} von \dependent{Maria} \\
\dep{nmod:poss} \ulink{https://universaldependencies.org/u/dep/nmod-poss.html} & possessor & \dependent{Marias} \head{Buch} \\
\dep{appos} \udelink{https://universaldependencies.org/u/dep/appos.html}{https://universaldependencies.org/de/dep/appos.html} & appositional noun phrase & das \head{Landeskriminalamt} (\dependent{LKA}) \\
\midrule
\multicolumn{3}{l}{\textit{\head{Noun-like} \rightarr{} \dependent{predicate}}} \\
\dep{acl} \udelink{https://universaldependencies.org/u/dep/acl.html}{https://universaldependencies.org/de/dep/acl.html} & clausal modifier & \head{Versuche}, dies zu \dependent{tun} \\
\dep{acl:relcl} \udelink{https://universaldependencies.org/u/dep/acl-relcl.html}{https://universaldependencies.org/de/dep/acl-relcl.html} & relative clause modifying NP & \head{Versuche}, die ich \dependent{unternehme} \\
\midrule
\multicolumn{3}{l}{\textit{\head{Noun-like} \rightarr{} \dependent{\_\_\_\_}}} \\
\dep{det} \udelink{https://universaldependencies.org/u/dep/det.html}{https://universaldependencies.org/de/dep/det.html} & determiner & \dependent{das} \head{Buch} \\
\dep{det:poss} \udelink{https://universaldependencies.org/u/dep/det-poss.html}{https://universaldependencies.org/de/dep/det-poss.html} & possessive pronoun & \dependent{mein} \head{Buch} \\
\dep{case} \udelink{https://universaldependencies.org/u/dep/case.html}{https://universaldependencies.org/de/dep/case.html} & adposition, comparative word & \dependent{an} dem \head{Tisch} \\
\dep{amod} \udelink{https://universaldependencies.org/u/dep/amod.html}{https://universaldependencies.org/de/dep/amod.html} & adjective & ein \dependent{altes} \head{Haus} \\
\dep{nummod} \udelink{https://universaldependencies.org/u/dep/nummod.html}{https://universaldependencies.org/de/dep/nummod.html} & number & \dependent{vier} \head{Pferde} \\
\dep{flat} \udelink{https://universaldependencies.org/u/dep/flat.html}{https://universaldependencies.org/de/dep/flat.html} & multi-word proper noun, date & \head{König} \dependent{Ludwig} \dependent{II} \\
\midrule
\multicolumn{3}{l}{\textit{\head{\_\_\_\_} \rightarr{} \dependent{\_\_\_\_}}} \\
\dep{conj} \udelink{https://universaldependencies.org/u/dep/conj.html}{https://universaldependencies.org/de/dep/conj.html} & conjunct & \head{Anna} und \dependent{Berta} \\
\dep{cc} \udelink{https://universaldependencies.org/u/dep/cc.html}{https://universaldependencies.org/de/dep/cc.html} & coordinating conjunction & Anna \dependent{und} \head{Berta} \\
\dep{punct} \ulink{https://universaldependencies.org/u/dep/punct.html} & punctuation & \head{Geht} sie \dependent{?} \\
\dep{advmod} \udelink{https://universaldependencies.org/u/dep/advmod.html}{https://universaldependencies.org/de/dep/advmod.html} & adverb, nicht & Das ist \dependent{nicht} \head{viel} \\
\dep{root} \ulink{https://universaldependencies.org/u/dep/root.html} & root & Sie \dependent{läuft} \\
\dep{fixed} \udelink{https://universaldependencies.org/u/dep/fixed.html}{https://universaldependencies.org/de/dep/fixed.html} & fixed expression & \head{nach} \dependent{wie} \dependent{vor} \\
\dep{parataxis} \udelink{https://universaldependencies.org/u/dep/parataxis.html}{https://universaldependencies.org/de/dep/parataxis.html} & coequal clause & Es ist, \dependent{sage} ich, zu \head{warm} \\
\dep{compound} \udelink{https://universaldependencies.org/u/dep/compound.html}{https://universaldependencies.org/de/dep/compound.html} & split compound word & \dependent{Telefon} \head{Buch} \\
\dep{goeswith} \ulink{https://universaldependencies.org/u/dep/goeswith.html} & randomly split word & \head{wer} \dependent{den} \\
\dep{orphan} \udelink{https://universaldependencies.org/u/dep/orphan.html}{https://universaldependencies.org/de/dep/orphan.html} & ellipsis & Ich gehe raus und \head{du} \dependent{rein} \\
\dep{reparandum} \udelink{https://universaldependencies.org/u/dep/reparandum.html}{https://universaldependencies.org/de/dep/reparandum.html} & disfluency & Nach \dependent{re–}, nach \head{links} \\
\dep{list} \ulink{https://universaldependencies.org/u/dep/list.html} & entire sentence is a list &  \\
\dep{dep} \ulink{https://universaldependencies.org/u/dep/dep.html} & ungrammatical relation & \\
\bottomrule
\end{tabular}

\renewcommand*{\arraystretch}{1}
\newpage

\subsection{Notes on specific dependency relations}
\label{sec:specific-deprels}

\paragraph{\dep{fixed}} See~\S\ref{sec:fixed}.

\paragraph{\dep{iobj}}
If a verb takes two accusative objects, typically, one gets the label \dep{iobj}. 
As of the UD~v2 guidelines, this is the more recipient-like object (typically the more animate one), e.g., \textit{Sie lehrt ihn}\textsubscript{\dep{iobj}} \textit{die französische Sprache}\textsubscript{\dep{obj}}, 
or \textit{Dies kostet ihn}\textsubscript{\dep{iobj}} \textit{den Verstand}\textsubscript{\dep{obj}}.

\begin{warning}{}
There is a recent discussion on GitHub \href{https://github.com/UniversalDependencies/docs/issues/1162}{[\#1162]}; maybe the guidelines for German will change?
\end{warning}

\paragraph{\dep{xcomp}}
Some verbs (like \textit{jdn etw nennen} ``to call sb sth'', \textit{jdn etw taufen} ``to christen sb sth'') use \dep{xcomp} instead of \dep{iobj} since the two objects essentially refer to the same entity (the structure \textit{X verbs Y Z} can be boiled down to \textit{Y is Z}).
Here, the more recipient-like/animate object is the \dep{obj} and the name they receive is an \dep{xcomp}: \textit{Nina nennt ihren Hund}\textsubscript{\dep{obj}} \textit{Rubi}\textsubscript{\dep{xcomp}}.

\subsection{Difficult cases: Which clausal relation?}

\paragraph{\dep{Acl} or not?}
\begin{itemize}
    \item If the clause modifies a noun \rightarr{} {\acl} or {\aclrelcl}
    \item If the clause modifies another clause \rightarr{} one of the options below ({\advcl}, {\ccomp}, {\xcomp})
\end{itemize}

\paragraph{Adverbial or not? ({\advcl} vs. {\ccomp/\xcomp})}
\begin{itemize}
    \item If you just drop the clause, is the overall sentence still fully grammatical and not ``weird''? \rightarr{} {\advcl}
    \item If you would need to replace the clause with a pronoun / if the dictionary entry for the head verb contains an \textit{etwas/jemand} that refers to the clause \rightarr{} {\ccomp},{\xcomp}
    \item Adverbial clauses often relate to: time \textit{(nachdem, bevor, bis, während, seit, als),} place \textit{(wo} -- but \textit{wo} also gets used as a relative marker!), manner \textit{(indem, als ob),} cause \textit{(weil),} purpose \textit{(damit),} effect \textit{(sodass),} contrast \textit{(obwohl, auch wenn),} condition \textit{(wenn, falls),} manner \textit{(als ob).} More examples \href{https://de.wikipedia.org/wiki/Adverbialsatz}{here}.
\end{itemize}

\paragraph{\dep{Ccomp} or {\xcomp}?}
If the dependent clause has its own subject, it is a {\ccomp}. If its subject is that of the main clause, it is an {\xcomp}.
Additionally, {\xcomp} is used for complements that are predicates without full clause structures (examples via \href{https://universaldependencies.org/de/dep/xcomp.html}{UD documentation}):
\begin{itemize}

\item Er \head{ließ} alle Demonstranten \dependent{verhaften}.
\item Er \head{blieb} dort \dependent{stehen}. 
\item Ich \head{lerne} \dependent{tanzen}. 
\item Wir \head{machen} uns \dependent{selbstständig}. 
\item Ich \head{fühle} mich \dependent{gezwungen}, dies zu tun. 
\end{itemize}

\paragraph{Examples}
\begin{itemize}
    \item ``Auch für Probetermine \head{nimmt sie sich sehr viel Zeit} und zeichnet alles genaustens vor, \dependent{sodass man sich vorstellen kann, wie das Ergebnis ist}.'' (GSD)
    \begin{itemize}
        \item Can easily be dropped: ``Auch für Probetermine nimmt sie sich sehr viel Zeit und zeichnet alles genaustens vor.'' \rightarr{} {\advcl}
        \item Basic form: \textit{sich Zeit nehmen} \rightarr{} {\advcl}
    \end{itemize}
    \item ``Auch für Probetermine nimmt sie sich sehr viel Zeit und zeichnet alles genaustens vor, so dass man sich \head{vorstellen} kann, \dependent{wie das Ergebnis ist}.'' (GSD) 
    \begin{itemize}
        \item Cannot be dropped, needs to be replaced: ``... dass man sich \dependent{das} \head{vorstellen} kann.'' \rightarr{} not {\advcl}
        \item Basic form: \textit{sich etwas vorstellen} \rightarr{} not {\advcl}
        \item Subordinate clause has its own subject \textit{(das Ergebnis)} \rightarr{} {\ccomp}, not {\xcomp}
    \end{itemize}
    \item ``Aber mal \head{abwarten}, \dependent{was sich in näherer Zukunft abspielt}...'' (GSD)
basic form: etwas abwarten \rightarr{} not {\advcl}
    \begin{itemize}
        \item Can be dropped, but with the understanding that something is missing. A more typical shortened reformulation would be something like ``\dependent{Das} \head{warten} wir ab'' \rightarr{} not {\advcl}
        \item Subordinate clause has its own subject \textit{(was)} \rightarr{} {\ccomp}, not {\xcomp}
    \end{itemize}
    \item ``Alle am Wort Gottes interessierte Personen sind herzlich \head{eingelanden}, Gottesdienst mit Liedern, Gebeten und Predigten mit den Mitgliedern zu \dependent{feiern}.'' (GSD)
    \begin{itemize}
        \item Can be dropped, but with the understanding that something is missing. A more typical shortened reformulation would be something like ``Alle sind \dependent{dazu} \head{eingeladen}'' \rightarr{} not {\advcl}
        \item Basic form: \textit{jemanden \dependent{zu etwas} einladen}
        \item The dependent clause has the same subject as the main clause: \textit{alle ... Personen} \rightarr{} {\xcomp}
    \end{itemize}
    \item ``Bei unserem nächsten Aufenthalt auf Sylt werden wir ganz bestimmt wieder hier \dependent{essen} \head{gehen}!'' (GSD)
    \begin{itemize}
        \item Can't be dropped \rightarr{} not {\advcl}
        \item Same subject as \textit{gehen} \rightarr{} {\xcomp}
    \end{itemize}
\end{itemize}

\subsection{Difficult cases: Parataxis or apposition?}
\label{sec:parataxis-apposition}
\dep{Parataxis} is typically between two clauses.
The linked words are often predicates.
Occasionally, noun phrases are involved.
Typical cases:
\begin{itemize}
    \item Reported speech where the speech tag interrupts the quote (otherwise: \ccomp, see guidelines \href{https://universaldependencies.org/u/dep/ccomp.html#reported-speech}{here})
    \item Two sentences that could also be separated with a full stop
    \item Interjections
    \item Affiliation bylines
    \item Question tags: ``oder?'', ``nicht?''
\end{itemize}
\noindent
An \dep{appos}ition is between two noun phrases; the linked words are nouns or noun-like. Typically, the head and dependent refer to the same entity and could be swapped.

Examples:
\begin{itemize}
    \item ``Aber über die Freundlichkeit, Zuverlässlichkeit und Komptenz des gesamten Team kann man nur \head{eines} sagen -- \dependent{Perfektion}'' (GSD)
    \begin{itemize}
        \item Although you can't swap the two words here, you can replace \textit{eines} with \textit{Perfektion} \rightarr{} \appos
    \end{itemize}
    \item ``Auch das Servieren der Speisen \head{ging} auffallend schnell, also ich könnte nicht so schnell \dependent{kochen}.'' (GSD)
    \begin{itemize}
        \item Could be separated with a full stop \rightarr{} \parataxis
        \item Clauses, not nouns \rightarr{} \parataxis
    \end{itemize}
    \item das \head{Landeskriminalamt} (\dependent{LKA})
    \begin{itemize}
        \item Can easily be swapped: ``das LKA (Landeskriminalamt)'' \rightarr{} \appos
    \end{itemize}
    \item ``\head{Barbara} Plank (\dependent{LMU})''
    \begin{itemize}
        \item Cannot be swapped \rightarr{} \parataxis
    \end{itemize}
    \item ``Tatsächlich gibt es \head{Bestrebungen}, den Straßenverkehr sicherer zu \dependent{machen}."
    \begin{itemize}
        \item Head is a noun, dependent a clause \rightarr{} \acl
    \end{itemize}
\end{itemize}
\noindent
We also use {\appos} in cases like the following, where the second entity (dependent) encompasses the first one (head): \textit{I live in \head{Munich}, \dependent{Germany}.}

For more use cases of {\appos}, see \S\ref{sec:key-value} and \S\ref{sec:ein-haufen}.

\newpage
\section{Lemmas}
\label{sec:lemmas}

We add German-language lemmas to the \texttt{MISC} column (\texttt{GermanLemma=...}). 
If we cannot figure out what a word means, we use the lemma \texttt{<unknown>}.

\paragraph{Pronouns}
Lemmatization of pronouns is modelled after the annotations in the German treebanks.
Non-possessive pronouns are lemmatized (e.g., \textit{mir, mich} \rightarr{} \textit{ich}; \textit{ihr}\textsc{.dat} \rightarr{} \textit{sie}).
For possessive pronouns, we only adjust the suffix (e.g., \textit{meine} \rightarr{} \textit{mein}).
The reflexive \textit{sich} remains \textit{sich}.

\paragraph{Articles}
Articles keep their gender and number in lemma form, but not the case (the gender/number needs to match that of the Bavarian noun, not that of the German lemma).

\paragraph{Other notes}
\begin{itemize}
    \item Words that were erroneously split and have the dependency relation \dep{goeswith} aren't annotated with a lemma. The head of the \dep{goeswith} sequence receives the lemma of the full sequence.
    \item We keep \textit{nimma} \translation{not anymore} as one token and give it the German lemma \texttt{nicht mehr}.
    \item Multi-word token meta entries don't receive any lemma annotation.
    \item Transparent abbreviations of German words are lemmatized as the full word (e.g., \textit{bzw.} \rightarr{} \textit{beziehungsweise}).
    \item Words that both have a less common German cognate and a more common non-cognate are currently annotated with both (e.g., \textit{Burschn} \rightarr{} \textit{Bursche/Junge} \translation{boy, young man}).
    \item The auxiliary subjunctive form of \textit{doa} (\textit{i dad, i tarat,} etc.) receives the lemma \textit{tun}.
\end{itemize}

\newpage
\section{General and German-related annotation decisions}
\label{sec:general}

This section contains annotation decisions that are not specific to Bavarian, but would also apply to Standard German and related dialects.
The German treebanks currently handle some of these cases differently (e.g., \S\ref{sec:title-name}).

\subsection{Abbreviations}
We split abbreviations when it makes sense \textit{(z.B., u.A.)} and tag the components as the POS tags of the individual words.
In cases like \textit{z.B. = zum Beispiel} \translation{for example})\, we use the POS tags of the words' head subtokens and \textit{do not} further analyze the structure of \textit{zu+m}: \tag{z.}{adp} \tag{B.}{noun}. 
One-word abbreviations \textit{(bspw., sog., bzw.)} are simply tagged like we would tag the unabbreviated word form.
The period stays attached to the abbreviation.

\subsection{Additions to proper names}
\label{sec:title-name}
For titles \textit{(Frau Müller, König Ludwig)} and suffixes \textit{(Ludwig II, Max Mustermann~Sr.)} of person names, as well as for `titles' of administrative divisions \textit{(Gemeinde~X, Landkreis~Y)}, we use \flatt.
This is in line with the \href{https://universaldependencies.org/u/dep/flat.html}{UD guidelines for \flatt}, although in reality it is handled in various different ways by different treebanks \citep{schneider-zeldes-2021-mischievous}.
The German treebanks currently disagree on whether \flatt, \dep{flat:name} or \compound{} should be used.

\subsection{Adjectives used adverbially}
\label{sec:adj-or-adv}
Following the German treebanks, we tag adverbally used adjectives as \pos{adj} and use the relation {\advmod} for them: \leftdep{\tag{schnäi}{adj}}{advmod}{\tag{laffa}{verb}} \translation{to run fast}.
For guidance on distinguishing adjectives used this way from adverbs, we refer to the notes on discerning between the \pos{adjd} and \pos{adv} classes in the STTS guidelines (\citealp{stts}, pp.~\href{https://www.ims.uni-stuttgart.de/documents/ressourcen/lexika/tagsets/stts-1999.pdf#page=57}{56--58}).

\subsection{Anonymized names}
\label{sec:anonymized-names}
We replace usernames with USERNAME, which we tag as \pos{propn}.

\subsection{Comparatives}
We use \pos{sconj} for \textit{wie/als}.
If a noun phrase follows, \textit{wie/als} gets {\case}, and the following noun phrase is annotated as {\obl} if attached to a phrase and {\nmod} if attached to another noun phrase.
If a clause follows, \textit{wie/als} is a {\markk}er and the clause is an \advcl.

\paragraph{Additional notes} 
Related links if we want to revisit this decision: \href{https://universaldependencies.org/workgroups/comparatives.html}{UD working group on comparatives}; \href{https://github.com/UniversalDependencies/docs/issues/767}{GitHub issue regarding German}.

\subsection{Copula}
Per UD guidelines, only variations of \textit{sei} \translation{to be} can be annotated as copulas; \textit{bleim} \translation{to remain}, \textit{wean} \translation{to become} and similar verbs are treated as full verbs.
More details on copulas in UD can be found \href{https://universaldependencies.org/v2/copula.html#guidelines-for-udv2}{here}.

\subsection{Dates and names of months, weekdays \& holidays}
\marginnote{
\begin{dependency}
\begin{deptext}
    31.\&12.\&2000\\
    \pos{adj} \& \pos{adj} \& \pos{num}\\
\end{deptext}
\depedge[right=5pt, arc angle=30]{1}{2}{\flatt}
\depedge[arc angle=90]{1}{3}{\flatt}
\end{dependency}
}[-20pt]
We use \pos{adj} for ordinal numbers and \pos{num} for cardinal numbers. The relation between the day, month and year is \flatt.
We use \pos{noun} for months, weekdays and holidays, and \pos{adv} for derived adverbs \textit{sonntags} \translation{on Sundays}.

\paragraph{In other treebanks} For an overview and discussion of how dates are handled in UD, see \citet{zeman-2021-date}. The German treebanks disagree on whether to use \pos{noun} or \pos{propn} for months, weekdays and holidays, and on whether to use \pos{adv} or \pos{noun} for words like \textit{sonntags} when they are capitalized.

\subsection{Dative objects}
\label{sec:dative-objects}
We currently follow the German guidelines, which reserve \href{https://universaldependencies.org/de/dep/iobj.html}{\iobj} for the (rare) second accusative object and instead prescribe \href{https://universaldependencies.org/de/dep/obl-arg.html}{\oblarg} for dative objects.
Note that the \href{https://universaldependencies.org/nds/#core-arguments-oblique-arguments-and-adjuncts}{Low Saxon} and \href{https://universaldependencies.org/gsw/#core-arguments-oblique-arguments-and-adjuncts}{Swiss German documentation} instead suggest {\iobj} for dative objects.

For Bavarian, we need to keep in mind the partial case syncretism for masculine and neuter dative and accusative definite articles and pronouns (\citealp{zehetner1978kontrastive}; \citealp[p.~98--99]{merkle1993bairische}); just because something looks like a dative\fslash{}accusative at first glance doesn't have to mean it actually is:

{\begin{center}
\marginnote{\margintext{Case syncretism in definite articles and personal pronouns, based on \citet[pp.~85, 122]{merkle1993bairische}.}}[-38pt]
\begin{tabular}{@{}lllllll@{}}
\toprule
 & \multicolumn{2}{c}{Stressed} & \multicolumn{2}{c}{Unstressed} & \multicolumn{2}{c}{Pronoun} \\
 \cmidrule(lr){2-3} \cmidrule(lr){4-5} \cmidrule(lr){6-7}
 & \textsc{Dat} & \textsc{Acc} & \textsc{Dat} & \textsc{Acc} & \textsc{Dat} & \textsc{Acc} \\ 
 \cmidrule(lr){2-2} \cmidrule(lr){3-3} \cmidrule(lr){4-4} \cmidrule(lr){5-5} \cmidrule(lr){6-6} \cmidrule(lr){7-7}
\textsc{Masc} & dem/\datlikeacc{den} & \datlikeacc{den} & am/\datlikeacc{an} & \datlikeacc{an}\hspace{5mm} & \acclikedat{eahm} & \acclikedat{eahm}/eahn\\
\textsc{Neut} & dem/\datlikeacc{den} & dees & am/\datlikeacc{an} & as/s & \acclikedat{eahm} & es/dees\\
\textsc{Fem} & dera & de & da & d & ia/iara & sie/de\\ \bottomrule
\end{tabular}
\end{center}}

\subsection{Dummy names: \textit{Hast du X gesehen?}}
\label{sec:dummy-names}
If nominals get replaced with dummy sequences in sentences like \textit{Gib dem Buach „...“ 5 Sterndal} \translation{Give 5 stars to book ``...''} or \textit{Wann kimmtn der Film A?} \translation{When is movie A on?}, we tag dummy tokens like \textit{A}, \textit{X} or \textit{XZY} as \pos{x}. 
If an ellipsis is used as the placeholder for a name (and saying the sentence out loud would likely involve saying something like \textit{Punkt Punkt Punkt} \translation{dot dot dot} or \textit{hm-hm-hm}), we tag it as \pos{sym}.
Either way, if the placeholder is used as an apposition to a noun, we annotate it accordingly:

\begin{dependency}
\begin{deptext}
das \& Buch \& „ \& ... \& “ \\
\pos{det} \& \pos{noun} \& \pos{punct} \& \pos{sym} \& \pos{punct}\\
\end{deptext}
\depedge{2}{4}{\appos}
\depedge[left=2mm]{4}{3}{\punct}
\depedge{4}{5}{\punct}
\end{dependency}
\hspace{3em}
\begin{dependency}
\begin{deptext}
der \& Film \& A \\
\pos{det} \& \pos{noun} \& \pos{x}\\
\end{deptext}
\depedge{2}{3}{\appos}
\end{dependency}

\noindent
If a name is replaced with a token like {USERNAME}, we annotate it as if it were the original token~(\S\ref{sec:anonymized-names}).

\paragraph{Other treebanks}
HDT uses \pos{x} in phrases like \textit{Nutzer A} \translation{user A}, GSD uses \pos{propn} (\textit{eines Teams A} \translation{of a team A}), EWT uses \pos{noun} \textit{(Party B)}.
We did not find any examples corresponding to our \textit{book ``...''} in our reference treebanks.

\subsection{Erroneously split words}
We use {\compound} for composite words (e.g., compound nouns) split across word boundaries, and {\goeswith} for `randomly' split words.

\subsection{Fixed expressions}
\label{sec:fixed}
We use the {\fixed} dependency for the following expressions:
\begin{itemize}
    \item \textit{ein paar} \translation{a few}
    \item \textit{ein wenig, ein bisschen} \translation{a bit}
    \item \textit{und zwar} \translation{namely}
    \item \textit{mehr/weniger als/wie} \translation{more/less than}
    \item \textit{ein und derselbe} \translation{one and the same}
    \item \textit{bis zu}
\end{itemize}

\noindent
As of 2.17, constructions like \textit{durch des} \translation{due to} and \textit{fir des} \translation{considering that} are no longer annotated as fixed expressions~(\S\ref{sec:durch-des}).

\subsection{Modal particles}
Because the \href{https://universaldependencies.org/de/#tags}{German UD guidelines} reserve \pos{part} only for \textit{nicht} and \textit{zu,} we treat modal particles like adverbs (which is what the German treebanks mostly do as well).

\subsection{Multi-part conjunctions: \textit{sowohl ... als auch,} etc.}
We use the following POS tags for multi-part conjunctions:
\begin{itemize}
    \item \tag{sowohl}{cconj} \textit{...} \tag{als}{cconj} \tag{auch}{adv} \translation{both ... and}
    \item \tag{entweder}{cconj} \textit{...} \tag{oder}{cconj} \translation{either ... or}
    \item \tag{weder}{cconj} \textit{...} \tag{noch}{cconj} \translation{neither ... nor}
\end{itemize}
\noindent
and annotate the dependencies as follows:

\begin{dependency}
\begin{deptext}
entweder \& X \& oder \& Y\\
\end{deptext}
\depedge{2}{1}{cc}
\depedge[label style={below}]{4}{3}{\cc}
\depedge[edge start x offset=4pt]{4}{2}{\conj}
\depedge[edge below, label style={below}, hide label, edge style={white}]{3}{4}{\fixed} %
\end{dependency}
\hspace{3em}
\begin{dependency}
\begin{deptext}
sowohl \& X \& als \& auch \& Y\\
\end{deptext}
\depedge{2}{1}{\cc}
\depedge[edge start x offset=4pt, arc angle=40]{5}{3}{\cc}
\depedge[edge start x offset=4pt]{5}{2}{\conj}
\depedge[edge below, label style={below}]{3}{4}{\fixed}
\end{dependency}

\subsection{Numeric ranges}
\marginnote{
\begin{dependency}
\begin{deptext}
300 \&[6pt] -- \&[6pt] 500\\
\pos{num} \& \pos{adp} \& \pos{num}\\
\end{deptext}
\depedge{1}{3}{\nmod}
\depedge[arc angle=40]{3}{2}{\case}
\end{dependency}
}[-20pt]
We treat numeric ranges like we treat the ``full'' version (i.e., if you were to replace -- with the preposition \textit{bis} \translation{to}), following GUM and EWT.

\subsection{Parenthetical key:value remarks}
\label{sec:key-value}
We annotate cases like \textit{Minga (amtli: München)} \translation{Minga (officially: Munich)} as follows:

\begin{dependency}
\begin{deptext}
Minga \& ( \& amtli \& : \& München \& )\\
\end{deptext}
\depedge{1}{3}{\appos}
\depedge{3}{5}{\appos}
\depedge[edge below, label style={below}]{3}{2}{\punct}
\depedge[edge below, label style={below}, right=12pt]{3}{4}{\punct}
\depedge[edge below]{3}{6}{\punct}
\end{dependency}

\noindent
For other apposition types, see \S\ref{sec:parataxis-apposition}.

\subsection{Participles: adjectives or verbs?}
\label{sec:participles}
We follow the STTS guidelines for distinguishing adjectives from verbal participles (\citealp{stts}, pp.~\href{https://www.ims.uni-stuttgart.de/documents/ressourcen/lexika/tagsets/stts-1999.pdf#page=25}{24--26}).

\subsection{Prepositional objects}
Following the German UD guidelines and examples (\href{https://universaldependencies.org/de/dep/obl.html}{\obl}, \href{https://universaldependencies.org/de/dep/obl-arg.html}{\oblarg}), we use {\obl} for prepositional phrases regardless of how core- or adjunct-like the phrase is.

\subsection{Pronouns as determiners}
\label{sec:pronouns-determiners}
Possessive pronouns are considered to be determiners with the relation {\detposs}, per \href{https://universaldependencies.org/de/pos/DET.html}{UD guidelines}.

For cases like, \textit{uns Linguisten} \translation{us linguists}, \textit{du Schmeichler} \translation{you flatterer},  we tag the pronoun as \pos{pron} and, following the recommendation by \citet{hohn-2021-towards}, label the relation {\dett}.
See \S\ref{sec:postponed-adjectives} for an example.

\subsection{Time}
\label{sec:time}

Numbers are tagged as \pos{num} regardless of whether they are written as digits or as words.
The dependency for prepositional phrase with the time is \obl{} or \nmod, depending on whether it is attached to a predicate or noun.
For phrases like \textit{12 Uhr} \translation{12 o'clock}, we use \dep{nummod} for the number.

\noindent
\begin{dependency}
\begin{deptext}
eine \& Erinnerung \& um \& vier\\
\end{deptext}
\depedge{4}{3}{\case}
\depedge{2}{4}{\nmod}
\end{dependency}
\hfill
\begin{dependency}
\begin{deptext}
erinnere \& mich \& um \& vier\\
\end{deptext}
\depedge{4}{3}{\case}
\depedge{1}{4}{\obl}
\end{dependency}
\hfill
\begin{dependency}
\begin{deptext}
vier \& Uhr\\
\end{deptext}
\depedge{2}{1}{\dep{nummod}}
\end{dependency}

\noindent
\translation{a reminder at 4}
\hspace{7em}
\translation{remind me at four}
\hfill
\translation{four o'clock}\\

\noindent
We annotate more complex or unusual structures as follows:

\begin{dependency}
\begin{deptext}
Stei \& an \& Wegga \& fia \& fünfe \& heid \& auf \& Nacht\\
\end{deptext}
\depedge{5}{4}{\case}
\depedge[right=4mm]{5}{6}{\advmod}
\depedge{8}{7}{\case}
\depedge{5}{8}{\nmod}
\end{dependency}

\noindent
\translation{Set an alarm for five tonight (lit. today on night)}
\attribution{xSID \textit{de-ba-test} 56}

\begin{dependency}
\begin{deptext}
um \& 3 \& nammiddog\\
\pos{adp} \& \pos{num} \& \pos{noun}\\
\end{deptext}
\depedge{2}{1}{\case}
\depedge{2}{3}{\appos}
\end{dependency}

\noindent
\translation{at 3 PM (lit. at 3 afternoon)}
\attribution{xSID \textit{de-ba-test} 57}

\subsection{Titles of books/songs/etc.}
\label{sec:song-titles}
If the title is \textit{not} in Bavarian or a closely related language/dialect: treat the words as \pos{propn}s connected with \flatt.
Otherwise, use normal tags and dependencies. If the overall construction is a copular clause, we can use \dep{nsubj:outer} (\href{https://universaldependencies.org/u/dep/nsubj-outer.html}{\textsc{u}}, no cases like that in our treebank as of yet):

\begin{dependency}
\begin{deptext}
Der \& Titel \& ist \& „ \& Herr \& Gröttrup \& setzt \& sich \& hin \& “\\
\end{deptext}
\depedge[arc angle=40]{7}{2}{\dep{nsubj:outer}}
\depedge[arc angle=30]{7}{3}{\dep{cop}}
\depedge[edge below]{5}{6}{\flatt}
\depedge[edge below]{7}{5}{\nsubj}
\depedge{7}{8}{\dep{expl:pv}}
\depedge[edge below, label style=below]{7}{9}{\dep{compound:prt}}
\end{dependency}

\noindent
\translation{The title is ``Mr Gröttrup sits down''}\\

\noindent
If the sentence is non-copular, we treat the title like a nominal:

\begin{dependency}
\begin{deptext}
Die \& Kurzgeschichte \& heißt \& „ \& Herr \& Gröttrup \& setzt \& sich \& hin \& “\\
\end{deptext}
\depedge{3}{2}{\nsubj}
\depedge[arc angle=30]{3}{7}{\obj}
\end{dependency}

\noindent
\translation{The short story is called ``Mr Gröttrup sits down''}\\

\noindent
If the title is replaced with a placeholder like \textit{XYZ} or \textit{...}, see \S\ref{sec:dummy-names}.

\subsection{Truncated words}
\label{sec:truncated}
In a case like \textit{Wirtschofts-, Vakeas- und Kuituazentren} \translation{economic, traffic and cultural centres}, we treat \textit{Wirtschofts-} as the head that \textit{Vakeas-} and \textit{Kuituazentren} are connected to via \conj.

This is somewhat unsatisfactory in that we would otherwise analyze \mbox{\textit{-zentren}} as the head of the compounds, but it aligns much better with how conjunctions are treated in UD.
We also use this when the split-off morpheme technically belongs to a different part of speech: \textit{be- und entladen} \translation{to load and unload} is tagged as \pos{verb~cconj~verb} with \textit{be-} as the head.

\subsection{Typos}
\label{sec:typos}

We do not correct \href{https://universaldependencies.org/u/overview/typos.html}{typos} or punctuation errors, but we annotate typos as such.
We mostly use intuitions from German to decide whether words are incorrectly split/merged, but the general principle is that if the words clearly encode different parts of speech and entities, we should split them.
For common merging patterns, see~\S\ref{sec:space-after-no}.

We split incorrectly merged words, and annotate the first word(s) with the MISC features \textsf{CorrectSpaceAfter=Yes} and \textsf{SpaceAfter=No}.

For incorrectly split words, we make the first subword the head of the sequence (with the feature \textsf{Typo=Yes}), connecting the others with \dep{goeswith}.
The head receives the German lemma corresponding to the entire sequence~(\S\ref{sec:lemmas}).

There is no Bavarian orthography, so we aren't concerned with exactly how a word is spelled.
If we, however, were to encounter a word with an undeniable typo (e.g., transposed letters resulting in an implausible spelling), we would annotate it with \textsf{Typo=Yes}.

\subsection{\textit{Bitte}}
If \textit{bitte} is used to mean `you're welcome,' we tag it as \pos{intj}.
If it is used in a sentence like \textit{Komm bitte mal her} \translation{Please come over here}, we consider it an \pos{adv}erb.
Note that it can also be an inflected verb or a noun.

\subsection{\textit{Durch/fir des, dass...} (instead of \textit{dadurch/dafür, dass...})}
\label{sec:durch-des}

We sometimes encounter constructions like \textit{durch des, dass...} \translation{due to} that correspond to a German pronominal adverb construction with \textit{da-} like \textit{dadurch, dass...}.
The latter is discussed in a GitHub issue \href{https://github.com/UniversalDependencies/docs/issues/1173}{[\#1173]}.

We now (2.17) annotate them like this:

\begin{dependency}
\begin{deptext}
    duach \& des \&, \& dass \& a \& arwat \& , \& vadejnt \& a \& ...\\
    \pos{adp} \& \pos{pron} \& \& \pos{sconj} \& \& \& \& \&\\
\end{deptext}
\depedge[]{2}{1}{\case} %
\depedge[]{2}{6}{\ccomp}
\depedge[]{8}{2}{\obl}
\end{dependency}

\noindent
\translation{Because he works, he earns ...}

\newpage
\subsection{\textit{Ein Haufen} \pos{noun}, \textit{eine Menge} \pos{noun}}
\label{sec:ein-haufen}
\marginnote{
\begin{dependency}
\begin{deptext}
ein \& Haufen \& Schrott\\
\end{deptext}
\depedge{2}{3}{\appos}
\end{dependency}
}[-20pt]
In structures like \textit{ein Haufen Formulare, eine Menge Formulare} \translation{a ton/lot of garbage}, we follow the German treebanks and connect the nouns with an {\appos}ition.

\begin{warning}{}
    This might need to be updated, based on a very recent GitHub discussion: \href{https://github.com/UniversalDependencies/docs/issues/1171}{[\#1171]}.
\end{warning}

\subsection{\textit{Gar nicht}}
We use \leftdep{\tag{gar}{adv}}{\advmod}{\tag{nicht}{part}} \translation{not at all}.
The German treebanks disagree on how to annotate this phrase (attach \textit{gar} to \textit{nicht} or to \textit{nicht}'s head).

\subsection{\textit{Selber/selbst}}
We attach {selber/selbst} \translation{him-/her-/themself} to the preceding noun if it is an adnominal construction; otherwise we attach it to the clause.
The relation is {\advmod} either way.
\citet[p.~136]{hole-2002-selbst} provides more details regarding the distinction between \textit{selbst} as an adnominal or adverbial intensifier, the examples are from his paper:
\begin{itemize}
    \item Adnominal: Der \head{Koch} \dependent{selbst} hat die Blaubeeren gepflückt. \translation{The cook himself picked the blueberries.}
    \item Adverbial: Der Koch hat die Blaubeeren \dependent{selbst} \head{gepflückt}. \translation{The cook picked the blueberries himself.}
\end{itemize}
\noindent
\textit{Selber} is a \pos{det}erminer per the \href{https://universaldependencies.org/de/pos/DET.html}{German guidelines}, and the German treebanks agree that \textit{selbst} is an \pos{adv}erb.

\subsection{\textit{So ein} \pos{adj noun}}
In some sentences, the adverb modifying an adjective can be placed in multiple positions:

\begin{itemize}
    \item \textit{Das ist so ein schönes Buch.} \translation{This is such a nice book.}
    \item \textit{Das ist ein so schönes Buch.} (non-crossing)
\end{itemize}
\noindent
We allow crossing dependencies, since \textit{so} modifies the adjective either way.
See \S\ref{sec:ein-ganz-ein} for a related Bavarian-specific phenomenon \textit{ein so ein schönes Buch}.

\subsection{\textit{Viel}}
\begin{itemize}
    \item \tag{so}{adv} + \tag{viel}{det} + \pos{noun}: \textit{so viel Musik, so viel Beifall} \translation{so much music, so much applause}
    \item \tag{so}{adv} + \tag{viel}{det} + \tag{wie}{sconj} \translation{as much as}
    \item \tag{viel}{adv} + \pos{adj}: \textit{viel später, viel wert} \translation{much later, worth a lot}
    \item \tag{viel}{adv} + \pos{verb}: \pos{adv} since it modifies a verb
    \item \tag{viel(e)}{det} + \pos{noun}: \textit{viele Babys, viel Gutes} \translation{many babies, a lot of good}
    \item \tag{viel(e)}{det} + \tag{von}{adp} + \pos{det} + \pos{noun}: \textit{viele von den Beteiligten, viel von der Stadt} \translation{many of those involved, much of the city}
    \item \tag{viel(e)}{det} + \pos{adj} + \pos{noun}: \textit{viele schöne Bilder} \translation{many pretty pictures}
    \item \tag{viel(e)}{det} without any noun (in accordance with the German treebanks): \textit{viele sind zu der Feier gekommen} \translation{many came to the party}
\end{itemize}

\newpage
\section{Bavarian-specific annotation decisions}
\label{sec:bavarian}

Although Bavarian is closely related to Standard German, there are some morphosyntactic differences.
In the following, we show examples for these as they occur in our data and explain how to annotate such structures.

\subsection{Noun phrase}
\label{sec:noun-phrase}

\subsubsection{Order of determiner and adverb \textit{(a ganz a...)}} 
\label{sec:ein-ganz-ein}
\marginnote{
\begin{dependency}
\begin{deptext}
\pos{det} \&[3pt] \pos{adv} \&[3pt] \pos{adj} \&[3pt] \pos{noun} \\
\end{deptext}
\depedge[edge start x offset=6pt]{4}{1}{\dett}
\depedge[left=3mm]{3}{2}{\advmod}
\depedge[edge start x offset=6pt, left=2mm]{4}{3}{\amod}
\end{dependency}
}[-33pt]
In German, if an adverb modifies an adjective in a noun phrase, the adverb appears between the determiner and the adjective (see margin).

For a small set of Bavarian intensifiers, alternative orders are possible (typically when the determiner is indefinite): the order of adverb and determiner can be reversed (\pos{adv~det~adj~noun}) and the determiner can be doubled (\pos{det~adv~det~adj~noun}; \citealp{lenz2014dynamik}; \citealp[pp.~89--90, 158]{merkle1993bairische}).
In such cases, we allow non-projective dependencies:
\enlargethispage{\baselineskip}

\begin{dependency}
\begin{deptext}
Frier wor des \& gonz \&a \&normales \&Wort \\
\gloss{Previously, it was} \& \gloss{very} \& \gloss{a} \& \gloss{normal} \& \gloss{word}\\
\& \pos{adv} \& \pos{det} \& \pos{adj} \& \pos{noun} \\
\end{deptext}
\depedge{5}{3}{\dett}
\depedge[left=2mm]{4}{2}{\advmod}
\depedge[edge style={gray}, label style={text=gray,below}]{5}{4}{\amod}
\end{dependency}

\noindent
\translation{It used to be a completely normal word.}\attribution{Wiki \textit{Walsch} \translation{Italian/Romance}}

\begin{dependency}
\begin{deptext}
In da englischn is \&[-6pt] a \& ganz \& a \& bläds \& Buidl \& drin \\
\gloss{In the English one is} \& \gloss{a} \& \gloss{very} \& \gloss{a} \& \gloss{silly} \& \gloss{picture} \& \gloss{inside} \\
\& \pos{det} \& \pos{adv} \& \pos{det} \& \pos{adj} \& \pos{noun} \\
\end{deptext}
\depedge{6}{2}{\dett}
\depedge{6}{4}{\dett}
\depedge{5}{3}{\advmod}
\depedge[edge style={gray}, label style={text=gray,below}]{6}{5}{\amod}
\end{dependency}

\noindent
\translation{The English [wiki] contains a very silly picture [...]}\\
\attribution{Wiki discussion \textit{Ottoman} \translation{sofa}}

\subsubsection{Personal names} 
In Bavarian, personal names are preceded by a determiner matching in case and gender \citep[pp.~69--70]{weiss1998syntax}, and the family name is often put before the given name \citep[p.~71]{weiss1998syntax}.
Following the general UD guidelines, we connect the parts of the name via a \textit{flat} relation: 

\begin{dependency}
\begin{deptext}
weder \& da\&  Schmidt\&  Bäda \& no \& d’\& Braun \& Maria\\
\gloss{neither} \& \gloss{the}.\textsc{m} \& \gloss{Smith} \& \gloss{Peter} \& \gloss{nor} \& \gloss{the}.\textsc{f} \& Brown \& Mary\\
\pos{cconj} \& \pos{det} \& \pos{propn} \& \pos{propn} \& \pos{cconj} \& \pos{det} \& \pos{propn} \& \pos{propn} \\
\end{deptext}
\depedge{3}{2}{\dett}
\depedge{3}{4}{\flatt}
\depedge{7}{6}{\dett}
\depedge{7}{8}{\flatt}
\end{dependency}

\noindent
\translation{neither Peter Smith nor Mary Brown [...]}\attribution{Cairo CICLing~12}

\subsubsection{Possession} 
Bavarian, like many German dialects and colloquial variants, eschews the genitive in favour of analytic possessive constructions \citep{fleischer2019syntax, buelow2021structures}.
One example is the prenominal dative construction, in which we analyze the possessor as an \textit{nmod}:

\begin{dependency}
\begin{deptext}
ohn \& in \& Lutha \& seina \& Iwasezung \\
\gloss{without} \& \gloss{the}.\textsc{dat} \& \gloss{Luther} \& \gloss{his} \& \gloss{translation} \\
\pos{adp} \& \pos{det} \& \pos{propn}\& \pos{det}\& \pos{noun}\\
\end{deptext}
\depedge{3}{2}{\dett}
\depedge[edge start x offset=3pt, edge end x offset=-3pt, label style={below}, below=1mm]{5}{4}{\detposs}
\depedge[edge start x offset=3pt]{5}{3}{\nmod}
\end{dependency}

\noindent
\translation{[...] without Luther's translation [...]}\attribution{Wiki discussion \textit{Ödenburg} \translation{Sopron}}

Alternatively, possession can be expressed with a prepositional phrase (common in colloquial German, and entirely parallel to the English \textit{X of Y} construction):

\begin{dependency}
\begin{deptext}
des \& Vaschwinden \& vo\& m \& Schädl \\
\gloss{the} \& \gloss{disappearance} \& \gloss{of} \& \gloss{the} \& \gloss{skull} \\
\pos{det} \& \pos{noun} \& \pos{adp}\& \pos{det}\& \pos{noun}\\
\end{deptext}
\depedge{2}{1}{\dett}
\depedge[label style={below}]{5}{4}{\dett}
\depedge{5}{3}{\case}
\depedge{2}{5}{\nmod}
\end{dependency}

\noindent
\translation{[...] the disappearance of the skull [...]}\\
\attribution{Wiki \textit{Sauschädelstöhln} \translation{Stealing pig's heads (custom)}}

\subsubsection{Postponed adjectives}
\label{sec:postponed-adjectives}
For emphasis (and especially when voicing annoyance), phrases of the pattern \pos{(adp) det adj noun} can be rearranged into \pos{(adp) det noun (adp) det adj} \citep[p.~168]{merkle1993bairische}.
We consider the postponed adjective to be an apposition of the noun.
In the following sentence in our corpus (pardon our Bavarian), \textit{du bleda Depp} `you stupid idiot' is re-arranged:

\begin{dependency}
\begin{deptext}
Hau di üba d'Heisa, \& du \& Depp  \& du \& bleda \& ! \\
\gloss{Get lost,}\& \gloss{you} \& \gloss{idiot} \& \gloss{you} \& \gloss{stupid} \& \gloss{!}\\
\& \pos{pron} \& \pos{noun} \& \pos{pron} \& \pos{adj} \\
\end{deptext}
\depedge{1}{3}{\vocative}
\depedge{3}{2}{\dett}
\depedge[arc angle=90]{3}{5}{\appos}
\depedge[left=2mm]{5}{4}{\dett}
\end{dependency}

\noindent
\translation{Get lost [lit.\ scram over the houses], you stupid idiot!}\attribution{Tatoeba 5657152}

\subsection{Verbs}
\label{sec:verb-phrase}

\subsubsection{Auxiliary \textit{tua}}

In addition to the auxiliary verbs named in the German guidelines, we include \textit{tua/doa} \translation{do}, which is used in several periphrastic constructions in conjunction with a lexical verb, both in indicative and subjunctive constructions \citep[pp.~65--67]{merkle1993bairische}.

\begin{dependency}
\begin{deptext}
Waun i du wa, \& tarat \& i \& 'n \& frogn \&.\\
\gloss{If I were you,} \& \gloss{do}.\textsc{1sg.sbjv} \& \gloss{I} \& \gloss{him} \& \gloss{ask} \& .\\
\& \pos{aux} \& \pos{pron} \& \pos{pron} \& \pos{verb} \& \\
\end{deptext}
\depedge[label style=below, arc angle=30]{5}{2}{\aux}
\end{dependency}

\noindent
\translation{If I were you, I would ask him.}\attribution{Tatoeba 5166978}

\subsubsection{Infinitives with \textit{z(u)}}
\label{sec:infinitive}
In German, many infinitive constructions require the marker \tag{zu}{part}.
In Bavarian, two similar constructions appear: one where a cliticized form of the marker \textit{(z)} is followed by a verbal infinitive,
and one where the marker is combined with a cliticized dative determiner (\textit{zum} or \textit{zun}) and a nominalized infinitive \citep{bayer1993zum, bayer2004klitisiertes-zu}.
In both cases, we annotate \tag{z(u)}{part} with {\markk} (as in the German treebanks), and in the latter, we separately annotate \tag{m/n}{det} with {\dett}:

\begin{dependency}
\begin{deptext}
{Ludwig van Beethoven hod de Gwohnheit ghobt,}\\
{\gloss{Ludwig van Beethoven had had the habit}}\\
\end{deptext}
\end{dependency}

\begin{dependency}
\begin{deptext}
genau 60 Kafääbaunan \& zu\& m \& oozöön\& , \\
\gloss{exactly 60 coffee beans} \& \textsc{inf} \& \gloss{the} \& \gloss{count} \& ,\\
\& \pos{part}\& \pos{det}\& \pos{noun}\&\&\\
\end{deptext}
\depedge[]{4}{2}{\markk}
\depedge[label style={below}]{4}{3}{\dett}
\end{dependency}

\begin{dependency}
\begin{deptext}
 um \& si \& draus\&  a Schalal Mokka\& z \& mochn \& . \\
\gloss{so as to} \& \textsc{refl} \& \gloss{out of it} \&  \gloss{a cup of coffee} \& \textsc{inf} \& \gloss{make} \& .\\
\&\&\&\& \pos{part}\& \pos{verb}\\
\end{deptext}
\depedge[right=2mm]{6}{5}{\markk}
\end{dependency}

\noindent
\translation{Ludwig van Beethoven had a habit of counting exactly 60 coffee beans in order to brew a cup of coffee from them} \attribution{Wiki \textit{Kafää} \translation{Coffee}}

\subsection{Pronouns and inflection}
See also \S\ref{sec:space-after-no} and \S\ref{sec:typos} for general guidelines (when the pronoun is clearly its own entity, treat it as a token: \textit{gibts}~\rightarr{}~\tag{gibt}{verb}~\tag{s}{pron}).

\subsubsection{Complementizer agreement \textit{(dassd, weilds, ...)}}
\label{sec:complementizer-agreement}
In Bavarian, reduced forms of second person (and, optionally, \textsc{1pl}) pronouns are used when they appear in the Wackernagel position immediately after complementizers \textit{(--sd}~\textsc{2sg,} \textit{--ds}~\textsc{2pl,} \textit{--ma}~\textsc{1pl)}.
These reduced forms are immediately attached to the previous word and can still be followed by a full pronoun for additional stress \citep[p.~119]{weiss1998syntax}:
\begin{table}[h]
\begin{tabular}{@{}llll@{}}
\toprule
 & Reduced pronoun & Full pronoun & Reduced + full \\ \midrule
\textsc{1sg} & wenn \clitic{e/i} gäh & wenn \fullpron{i} gäh (?) & --- \\
\textsc{2sg} & wenn\clitic{sd} af Minga kimmsd & --- & wenn\clitic{sd} \fullpron{du} af Minga kimmsd \\
\textsc{3sg} & wenn \clitic{a} des duad & wenn \fullpron{ea} des duad & --- \\
\textsc{1pl} & wem \clitic{ma} af Minga fahrn & wenn \fullpron{mia} af Minga fahrn & (wem\clitic{ma} \fullpron{mia} af Minga fahrn) \\
\textsc{2pl} & wenn\clitic{ds} af Minga kemds & --- & wenn\clitic{ds} \fullpron{ees} af Minga kemds \\ 
\textsc{3pl} & wenn \clitic{s} genga & wenn \fullpron{se} genga (?) & --- \\
\bottomrule
\end{tabular}
\caption*{Pronoun forms after complementizers. Our tokenization is indicated by whitespace. Adapted from \citet[pp.~119, 126 -- \textsc{1sg} reduced, \textsc{2sg, 3sg, 1pl, 2pl}]{weiss1998syntax} and \citet[p.~189 -- \textsc{1sg} reduced, \textsc{3pl} reduced]{merkle1993bairische}; the entries marked with (?) are not in either source but extrapolated by the guideline authors. \translation{When I/they go; When you.\textsc{sg/pl} come to Munich; If he does that; When we go to Munich.}}
\end{table}

\noindent
Whether these constructions should be analyzed as a word followed by an enclitic pronoun or as inflected complementizers is debatable (for an overview of the different arguments, see \citealp[pp.~123--133]{weiss1998syntax}).
To what extent \textsc{1pl} should be included in this analysis depends on the dialect and linguist \citep[cf.][p.~123, fn.~48]{weiss1998syntax}.

For our annotations, we follow \citet{bayer2013klitisierung} and adopt the interpretation of inflection for the second person (and for \textit{doubly marked} \textsc{1pl} cases):

\noindent
\begin{dependency}
\begin{deptext}
Er wüll, \& das'st \& Du \& redst \& .\\
\gloss{He wants} \& \gloss{that}.\textsc{2sg} \& \gloss{you}.\textsc{sg} \& \gloss{talk}.\textsc{2sg} \& .\\
\& \pos{sconj} \& \pos{pron} \& \pos{verb} \& \\
\end{deptext}
\depedge[label style=below]{4}{2}{\markk}
\depedge[label style=below]{4}{3}{\nsubj}
\end{dependency}
\hfill
\begin{dependency}
\begin{deptext}
Er wüll, \& das'st \& \& redst \& .\\
\gloss{He wants} \& \gloss{that}.\textsc{2sg} \& \& \gloss{talk}.\textsc{2sg} \& .\\
\& \pos{sconj} \& \& \pos{verb} \& \\
\end{deptext}
\depedge[label style=below]{4}{2}{\markk}
\end{dependency}

\noindent
\translation{He wants you to talk.}\attribution{Wiki \textit{Konjunktiona} \translation{Conjunctions}}

\begin{dependency}
\begin{deptext}
Er wüll, \& das \& i \& redt \& .\\
\gloss{He wants} \& \gloss{that} \& \gloss{I} \& \gloss{talk}.\textsc{1sg} \& .\\
\& \pos{sconj} \& \pos{pron} \& \pos{verb} \& \\
\end{deptext}
\depedge{4}{2}{\markk}
\depedge[label style=below]{4}{3}{\nsubj}
\end{dependency}
\hfill
\begin{minipage}{0.3\textwidth}
\vspace{-3\baselineskip}
\textcolor{gray}{(No version with\\ dropped \textit{i} possible.)}
\end{minipage}

\noindent
\translation{He wants me to talk.}\attribution{Wiki \textit{Konjunktiona} \translation{Conjunctions}}
\\

\noindent
The endings \textit{-sd} and \textit{-ds} can also be attached to other words \citep[pp.~127--128]{merkle1993bairische}; see the following examples (ibid.):
\begin{itemize}
    \item \textit{dees Bia, des wo\clitic{ds} neilich drungga habds} \translation{the beer that\textsc{.2pl} you drank the other day}
    \item \textit{i wui wissn, wea\clitic{sd} du bisd} \translation{I want to know who\textsc{.2sg} you are}
    \item \textit{Du soisd sång, an wäichan Schuah\clitic{sd} wuisd.} \translation{You have to say which shoe\textsc{.2sg} you want}
    \item \textit{wia schnäi\clitic{sd} fahsd} \translation{how fast\textsc{.2sg} you go} -- this is often replaced with a \textit{dass} construction~(\S\ref{sec:dass}): \textit{wia schnäi das\clitic{sd} fahsd}
\end{itemize}
\noindent
If you encounter any such cases, please bring them up during a meeting.

\begin{warning}{}
There are a few cases where people write, e.g., \textit{dass d} with a blank space in between. We solve this with {\goeswith}.
\end{warning}

\subsubsection{\textsc{1pl} \textit{-ma}}
\label{sec:mia-samma}

\paragraph{Double-marking (\textit{mia gemma})}

The \textsc{1pl.pres} inflection of verbs is typically identical to the infinitive form: \textit{mia genga} \translation{we go}.
However, it is also possible to add \textit{-ma} to the stem of the verb instead: \textit{mia gemma} \citep[p.~127]{merkle1993bairische}. (In this example, the nasal of ending of the stem, \textit{-ng}, assimilated to the \textit{m-}).
Although this ending historically comes from a cliticized form of the pronoun, we simply analyze it as inflection: \tag{mia}{pron} \tag{gemma}{verb}.

Whether to treat \textit{-ma} as inflectional morpheme or clitic is controversial \citep[p.~123, fn.~48]{weiss1998syntax}.
However, this annotation decision is consistent\marginnote{\margintext{The same applies to the \textit{-t} in the Standard German \textsc{2sg} ending \textit{-st} \citep[p.~127]{weiss1998syntax}.}} with how we annotate the 2nd person inflection of, e.g., \textit{du gähsd} \translation{you\textsc{.sg} go} and \textit{ees gähdds} \translation{you\textsc{.pl} go}, although \textit{-d} and \textit{-s} also evolved from pronouns -- a fact that Bavarian speakers are likely not aware of \citep[p.~127]{weiss1998syntax}.
This decision also lends itself well to UD annotation: it is unclear what dependency label \textit{-ma} should get, since the independent pronoun \textit{mia} is already the \nsubj.

\paragraph{Only \textit{-ma}, no other \nsubj}
While we currently do not have any occurrences of this, Bavarian allows clauses that only have \textit{-ma} and no other \nsubj{}: in VS clauses like questions (e.g., \textit{Hamma des?} \translation{Do we have it?}) and in imperatives (\textit{Gemma!} \translation{Let's go!}). 
If we come across these, we will have to decide how to treat them.
The following seems sensible:

\paragraph{Imperatives \textit{(gemma!)}}
\citet[pp.~253--254]{bayer1984comp} argues that the imperative should be treated as inflection since no version without \textit{-ma} exists \textit{(*Genga mia!)}.

\paragraph{VS clauses \textit{(gemma?)}}
The situation in verb-first clauses appears to be controversial (inflection or clitic?), especially as dialects in Lower Bavaria seem to differ from other Bavarian dialects with respect to \textit{-ma} (\citealp[p.~252]{bayer1984comp}, \citealp[p.~123, fn.~48]{weiss1998syntax}, \citealp[p.~201]{altmann1984system}). 
To keep our annotations straightforward and overall consistent, the most simple decision would be to split off \textit{-ma} and treat it as a \pos{pron} and \nsubj{}, unless there also is a separate \textit{mia} (see above for such cases).

\newpage
\subsubsection{Dropped 2nd person pronouns}
\label{sec:dropped-second}
\marginnote{
\margintext{Personal pronouns\\ before and after verbs, based on \citet[pp.~63--64]{merkle1993bairische}.}
\begin{tabular}{@{}l@{\hspace{4pt}}l@{}}
\toprule
SV & VS \\
\midrule
i håb & håw i/{a} \\
du håsd & håsd{\ \ } \\
ea/sie/es håd & håd {a/s} \\
mia ham & ham{ma} \\
ia habds & habds{\ \ } \\
de ham & ham {s} \\ \bottomrule
\end{tabular}

}
Second person pronouns can be omitted when they occur after a correspondingly inflected verb.
Consider the table in the margin and compare the following two sentences:

\noindent
\begin{dependency}
\begin{deptext}
Vo \& wos \& redst\&  Du\& ?\\
\gloss{of} \& \gloss{what} \& \gloss{talk}.\textsc{2sg} \& \gloss{you}.\textsc{sg} \& ?\\
\pos{adp} \& \pos{pron} \& \pos{verb} \& \pos{pron} \& \\
\end{deptext}
\depedge{3}{4}{\nsubj}
\depedge[edge style={gray}, label style={text=gray}]{2}{1}{\case}
\depedge[edge style={gray}, label style={text=gray}]{3}{2}{\obl}
\end{dependency}
\hfill
\begin{dependency}
\begin{deptext}
Kaunst \& aufstehn \& ?\\
\gloss{Can}.\textsc{2sg} \& \gloss{get up}.\textsc{inf} \& ?\\
\pos{aux} \& \pos{verb} \& \\
\end{deptext}
\depedge[edge style={gray}, label style={text=gray}]{2}{1}{\aux}
\end{dependency}\hspace{1em}\phantom{.}

\noindent
\translation{What are you talking about?}\hfill\translation{Can you get up?}\hspace{2.8em}\phantom{.}\\
(Wiki \textit{Konjunktiona} \translation{Conjunctions})\hfill(Tatoeba 10673747c)\hspace{1em}\phantom{.}
\\

\noindent
This simply means that some sentences won't have an \nsubj.

We do \textit{not} analyze this as an inflection ending \textit{-s(d)} followed by or merged with a reduced pronoun \textit{d} (or \textit{-d(s)} and \textit{s} for \textsc{2pl}) -- while this would be an etymological analysis of \textit{-sd/ds}, it is unlikely speakers think of it that way \citep[p.~127]{weiss1998syntax}.

\subsubsection{Dropped \textit{es} after \textit{-s}}
\label{sec:dropped-es}

In a few sentences in our treebank, the pronoun \textit{(e)s} \translation{it} is dropped after (or merged with?) \textit{-s} (e.g., \textit{Is heid bewölkt?} \translation{Is it cloudy today}; xSID \textit{de-ba-test} 18).
If a merge is indicated orthographically (e.g., the \textit{iss} in \textit{Im Nordboarischn dageeng iss mèjer wej im Standarddaitschn.} \translation{In North Bavarian however, it is more like in Standard German.}; Wiki discussion \textit{Boarische Umschrift} \translation{Bavarian transcription}), we separate the sequence into two tokens (\textit{is} and \textit{s}). 
Otherwise, we leave the token as is -- the sentence then lacks an \nsubj{} (see also \S\ref{sec:dropped-second}).

\subsection{Other annotation decisions for Bavarian}
\label{sec:bavarian-other}

\subsubsection{Additional complementizer \textit{dass}}
\label{sec:dass}
The adverb, relative pronoun, or question word introducing a subordinate clause can be followed by an additional conjunction \textit{dass} `that'
(\citealp[pp.~29--30]{weiss1998syntax}; \citealp[pp.~190--191]{merkle1993bairische}), which we consider a {\markk}er:

\begin{dependency}
\begin{deptext}
Jezz mechad i owa wissn, \&[-3pt] wia \& lang \&[-2pt] das \&[-3pt] des \&[-3pt] no \& dauat \&[-4pt] .\\
\gloss{Now I'd like to know} \& \gloss{how} \& \gloss{long} \& \gloss{that} \& \gloss{this} \& \gloss{still} \& \gloss{takes} \& .\\
\& \pos{adv} \& \pos{adj}\& \pos{sconj}\& \pos{pron}\& \pos{adv}\& \pos{verb} \&\\
\end{deptext}
\depedge[left=2mm]{3}{2}{\advmod}
\depedge{7}{3}{\advmod}
\depedge[label style=below]{7}{4}{\markk}
\end{dependency}

\noindent
\translation{Now I'd like to know how long this will still take.}\\\attribution{Wiki \textit{Pronomen} \translation{Pronouns}}

\subsubsection{Interjections}
\label{sec:intj}
Some interjections evolved from words originally belonging to other parts of speech, but we annotate them as \pos{intj}.
This includes \textit{gäi/gell} (and the optional polite version \textit{gäins} or \rightdep{\textit{gäin}}{fixed}{\textit{S}}), which is derived from inflected forms of what corresponds to the German verb \textit{gelten} \translation{to be valid} \citep[p.~197]{merkle1993bairische}, \textit{gäh/gö} \translationwithnote{lit.}{go\textsc{.imp}} when used as an interjection \citep[cf.][p.~76]{merkle1993bairische}, and \textit{mei} \translationwithnote{lit.}{my} \citep[cf.][p.~142]{merkle1993bairische}.
When \textit{sowas} is used as an interjection, we annotate it as such.

\subsubsection{Negative concord}
Unlike German, Bavarian allows for negative concord in constructions with (inflected forms of) \textit{koa} `no' \citep[pp.~167--168]{weiss1998syntax}:

\begin{dependency}
\begin{deptext}
Se \& hom \& koane \& Haxn \& ned \\
\gloss{They} \& \gloss{have} \& \gloss{no} \& \gloss{legs} \& \gloss{not} \\
\pos{pron} \& \pos{verb} \& \pos{det} \& \pos{noun} \& \pos{adv} \\
\end{deptext}
\depedge[edge style={gray}, label style={text=gray}]{2}{1}{\nsubj}
\depedge[edge style={gray}, left=2mm, label style={text=gray, below}]{2}{4}{\obj}
\depedge{2}{5}{\advmod}
\depedge[label style=below]{4}{3}{\dett}
\end{dependency}

\noindent
\translation{`They have no legs [...]'}\attribution{Wiki \textit{Fiisch} \translation{Fish}}

\subsubsection{Relative pronouns and particles}
Where German uses the relative pronouns \textit{der/die/das} `that, which', Bavarian can append the invariant relative marker \textit{wo} \citep{moser2023relative-particles}.
In some dialects, the relative marker is expressed as \textit{was} \citep{pittner1996attraktion}, and in our data we also find \textit{wie/wej} in northern regions.
We tag the relative pronoun as \pos{pron} (as in the German treebanks) and the relative marker as \pos{sconj} with the relation {\markk}:

\noindent
\hspace{-7pt}
\begin{dependency}
\begin{deptext}
'S gibt owa no vui \& Junge \&, \& de \&  wo \&  s' \& Boarische  \& no \& vastenga\\
\gloss{But there are still many}  \& \gloss{young ones}.\textsc{acc} \&, \& \textsc{rel}.\textsc{3pl.nom} \& \textsc{rel} \& \gloss{the} \& \gloss{Bavarian} \& \gloss{still} \& \gloss{understand} \\
\& \& \& \pos{pron} \& \pos{sconj} \& \pos{det} \&  \pos{noun} \&  \pos{adv} \&  \pos{verb} \\
\end{deptext}
\depedge[label style={below}, arc angle=30]{9}{5}{\markk}
\depedge[arc angle=30]{9}{4}{\nsubj}
\depedge[arc angle=30, edge style={gray}, label style={text=gray}]{2}{9}{\aclrelcl}
\end{dependency}

\noindent
\translation{However, there are still many young people who still understand Bavarian}\\\attribution{Wiki \textit{Minga} \translation{Munich}}

\noindent
In certain situations, the relative pronoun can be dropped in Bavarian if the relative marker is present \citep{pittner1996attraktion}. 
This can for instance happen when the case of the relative pronoun matches that of the modified noun, but also when the relative pronoun would be in the nominative case:

\noindent
\hspace{-7pt}
\begin{dependency}
\begin{deptext}
'S gibt owa no vui \& Junge \&, \&  wo \&  s' \& Boarische  \& no \& vastenga\\
\gloss{But there are still many}  \& \gloss{young ones}.\textsc{acc} \&, \& \textsc{rel} \& \gloss{the} \& \gloss{Bavarian} \& \gloss{still} \& \gloss{understand} \\
\& \& \& \pos{sconj} \& \pos{det} \&  \pos{noun} \&  \pos{adv} \&  \pos{verb} \\
\end{deptext}
\depedge[label style={below}, arc angle=30]{8}{4}{\markk}
\depedge[arc angle=30, edge style={gray}, label style={text=gray}]{2}{8}{\aclrelcl}
\end{dependency}

\begin{warning}{}
    \textit{Wo} can also appear as a location adverb introducing a subordinate clause: \textit{Der Ort, \tag{wo}{adv}{\_\advmod} ich aufgewachsen{\_\acl} bin.} \translation{The place where I grew up.} vs. \textit{Der Ort, (der) \tag{wo}{sconj}{\_\markk} gerade erwähnt{\_\aclrelcl} wurde.} \translation{The place that just got mentioned.}
\end{warning}

\paragraph{Additional notes}
The difference between a \textit{relative pronoun} and a \textit{relative marker} is that the pronoun might match the case\fslash{}number\fslash{}gender\fslash{}animateness/... of the modified noun, whereas the marker does not inflect.
English has both relative pronouns \textit{(who(m), which)} and a relative marker \textit{(that).}
\href{https://github.com/UniversalDependencies/docs/issues/223}{For reasons of practicality,} \textit{that} is however treated in UD as if it were a relative pronoun. %
However, in the same discussion, Chris Manning also writes:
\begin{quote}
    There is an argument from Middle English that \textit{that} should be analyzed as a complementizer while \textit{who} and \textit{which} are relative pronouns ... there are good arguments for this, including that you can get \textit{both}:\\
    Every word which that she of hire herde (Troilus and Criseyde, II.899)
\end{quote}
\noindent
This double relativization seems very similar to the situation in Bavarian.
How do treebanks in other languages deal with relative markers?
\begin{itemize}
    \item UZH considers \textit{wo} a pronoun. However, in Swiss German (unlike Bavarian), \textit{wo} is the sole relativizer \citep{moser2023relative-particles}.
    \item North Germanic languages use the invariant \textit{som} \textsc{(nor, swe, dan)}, \textit{sem} \textsc{(isl)} and \textit{sum} \textsc{(fao)} to mark relative clauses (on its own, with no relative pronouns). 
    The Norwegian and Icelandic treebanks treat it as \pos{sconj}, the Swedish and Danish ones as \pos{pron}, and the Faroese treebanks are split between the two options.
\end{itemize}

\subsubsection{Temporal expressions}
For expressions like \textit{um 3 Nachmittag}, see \S\ref{sec:time}.

\newpage
\section{Potential future updates}
\label{sec:future}

Some annotation decisions might need to be revisited, especially as the general and/or German-specific UD guidelines get refined.

\begin{itemize}
    \item Double-accusative verbs and \dep{iobj}~(\S\ref{sec:specific-deprels}): \href{https://github.com/UniversalDependencies/docs/issues/1162}{[\#1162]}
    \item \textit{ein Haufen Zeug, ein Glas Wein}~(\S\ref{sec:ein-haufen}): \href{https://github.com/UniversalDependencies/docs/issues/1171}{[\#1171]}
    \item Revisit the fixed expressions~(\S\ref{sec:fixed}): Are there any we can ``un-fix''? Add \textsf{ExtPos} for all fixed expressions to placate the \href{https://quest.ms.mff.cuni.cz/udvalidator/cgi-bin/unidep/validation-report.pl}{validator}?
\end{itemize}

\newpage
\addcontentsline{toc}{section}{References}
\bibliography{lib, ud}
\bibliographystyle{vbib}

\end{document}